# DragonFly: Single mmWave Radar 3D Localization of Highly Dynamic Tags in GPS-Denied Environments

Skanda Harisha, Jimmy G. D. Hester, Aline Eid
{skandah,jghester,alineeid}@umich.edu
University of Michigan, USA

## Abstract

The accurate localization and tracking of dynamic targets, such as equipment, people, vehicles, drones, robots, and the assets that they interact with in GPS-denied indoor environments is critical to enabling safe and efficient operations in the next generation of spatially-aware industrial facilities. This paper presents DragonFly, a 3D localization system of highly dynamic backscatter tags using a single MIMO mmWave radar. The system delivers the first demonstration of a mmWave backscatter system capable of exploiting the capabilities of MIMO radars for the 3D localization of mmID tags moving at high speeds and accelerations at long ranges by introducing a critical Doppler disambiguation algorithm and a fully-integrated cross-polarized dielectric-lens-based mmID tag consuming a mere 68 µW. DragonFly was extensively evaluated in static and dynamic configurations, including on a flying quadcopter, and benchmarked against multiple baselines, demonstrating its ability to track the positions of multiple tags with a median 3D accuracy of 12 cm at speeds and acceleration on the order of $10\,\mathrm{m\,s^{-1}}$ and $4\,\mathrm{m\,s^{-2}}$ and at ranges of up to 50 m.

## 1 Introduction

Localization services are a feature of modern day life. For most—especially those under 30, it is hard to imagine a world without the precise positioning delivered by smartphones and their connection to the Global Navigation Satellite System (GNSS). This ubiquitous capability has fueled new business models worth billions of dollars, powering companies like Uber, Lyft, Doordash, and Samsara. Yet, its greatest strength—global reliance on GNSS—is also its Achilles heel. The moment GNSS signals are lost or denied, localization services quickly collapse. Nowhere is this more evident than indoors, where signals are too attenuated to be useful. Even in cases where signals are faintly available, their raw accuracy already limited to the decameter range—degrades well beyond the needs of indoor localization.

There is a clear opportunity for indoor services especially in increasingly automated industrial, logistics, and retail facilities but it hinges on the availability of responsive, accurate, and cost effective localization technologies. Current state of the art Real Time Localization Systems (RTLS) include Bluetooth Angle of Arrival (e.g., Quuppa), Ultrasonic (Marvelmind Robotics, ZeroKey), WiFi (Deeyook), and Ultra Wideband (Ubisense, Decawave). Yet none of these support long battery life for mobile trackers, nor can they deliver 2D let alone 3D positions without aggregating data from multiple anchors or readers. These limitations make RTLS deployments prohibitively expensive to install and maintain. Notably, today's commercial leader, UWB, requires over 100 anchors per 100k sqft to enable trilateration, driving costs above $300k per industrial facility, including installation. Achieving long-range, single-anchor 3D localization could dramatically cut infrastructure density, hardware, and deployment costs, making such systems far more practical and widespread.

Backscatter modulation—the reflection of an impinging wireless signal via dynamic modification of a tag's Radar Cross-Section (RCS) is a promising solution to the first limitation. Yet, its link budget scales as $\frac{1}{R^4}$ in monostatic systems, where transmitter and receiver are collocated, restricting range and driving untenable reader densities. The rise of commodity mmWave systems offers a new path forward. Contrary to common belief, increasing the operating frequency of a backscatter system with fixed dimensions can extend maximum range [9]. When combined with retrodirective arrays [17] at mmWave frequencies [10], backscatter's core liability becomes an asset. The same reflection principle behind the $\frac{1}{R^4}$ scaling can be exploited by retrodirective structures such as Van Attas [9] and Rotman lenses [6], which passively and intensely focus reflections back toward the source—more than compensating for higher frequency path loss.

What, then, can be done to cost effectively localize and track such tags? The most practical answer lies in the most common and affordable mmWave radio: Frequency Modulated Continuous Wave (FMCW) radar. This approach stands out for combining wide scanning bandwidths with narrow baseband bandwidths, enabling both high range resolution and low processing loads. FMCW radars were first explored for backscatter tracking in the late 1990s [36] and later applied to localizing retrodirective tags [3, 8, 31]. Yet, even the latest and most mature systems [3, 5, 31] remain insufficient for future RTLS needs: some cannot achieve full 3D localization [5, 31], while others require multiple radars [3], as illustrated in Figure 1. Moreover, none can handle the dynamic environments that real world RTLS deployments demand. Unlocking these capabilities from low cost mmWave systems requires a fundamentally new approach—this is where DragonFly comes in.

This paper introduces DragonFly (shown in Figure 1), a mmWave backscatter localization system capable of tracking targets moving at accelerations up to $5\,\mathrm{m\,s^{-2}}$ in 3D and at ranges up to 50 m, even in dynamic environments. It achieves median accuracy better than 12 cm with latencies under 13.6 ms, using only a single MIMO time-divided FMCW radar. The custom-designed, fully integrated tag employs a lens architecture to provide $100° \times 100°$ azimuth and elevation coverage while consuming just 68 µW. Table 1 benchmarks DragonFly against state of the art backscatter localization systems, highlighting its unique ability to deliver precise, real-time, long-range localization of ultra low power dynamic tags.

DragonFly makes three key contributions, which firmly promote the advent of general, high performance, and low cost real time localization systems:

**First fully integrated lens based mmID:** Achieving long detection ranges with mmWave backscatter tags depends on engineering electrically large antenna systems that direct high reflection gain back to the reader. The most common approach is the Van Atta reflectarray, where antennas are paired symmetrically about an axis. However, as the array grows, the connecting lines between pairs become excessively long and lossy, limiting scalability beyond roughly three rows or columns [35]. An alternative is to use lenses for retrodirectivity, offering far greater scalability. In this work, we present the first fully integrated tag to employ this approach at an unprecedented scale.

**Acceleration robust modulation and localization:** State-of-the-art (SOTA) FMCW-based mmWave localization systems [3, 5, 31] aggregate many FMCW chirps to suppress clutter before detecting and localizing backscatter tags. However, this approach couples the Doppler and range dimensions, leading to poor detection of moving or accelerating tags. While complex modulation schemes can mitigate this issue [5], they add power and cost overheads. In contrast, we introduce a single frequency intra chirp modulation scheme, where just one FMCW chirp suffices to suppress clutter and simultaneously detect and localize tags in both radial and azimuth dimensions significantly alleviating the limitations of SOTA systems.

**Time-division transmission Doppler disambiguation for MIMO FMCW:** Multiple In Multiple Out operation is a common feature of a wide variety of wireless systems. Such a MIMO system utilizing N and M transmitting (Tx) and receiving (Rx) antennas (respectively) can approximate the spatial capabilities of a MISO system featuring 1 and $N \times M$ TX and RX antennas. Compared to the MISO approach, the MIMO system then requires N fewer receivers, ADCs, and data processing, making it significantly cheaper. It is, therefore, no surprise that commodity FMCW radars adopt this scheme. However, these N TX channels need to be somehow multiplexed. The most common approach for this is time division—all of the TX channels sequentially emit one at a time. While this can relatively easily be accommodated with static tags [2], movement introduces unknown phase changes from one time to the next (Doppler), thereby all but nullifying MIMO capabilities. DragonFly introduces the first scheme capable of disambiguating Doppler from TX channel switching for mmWave backscatter systems.

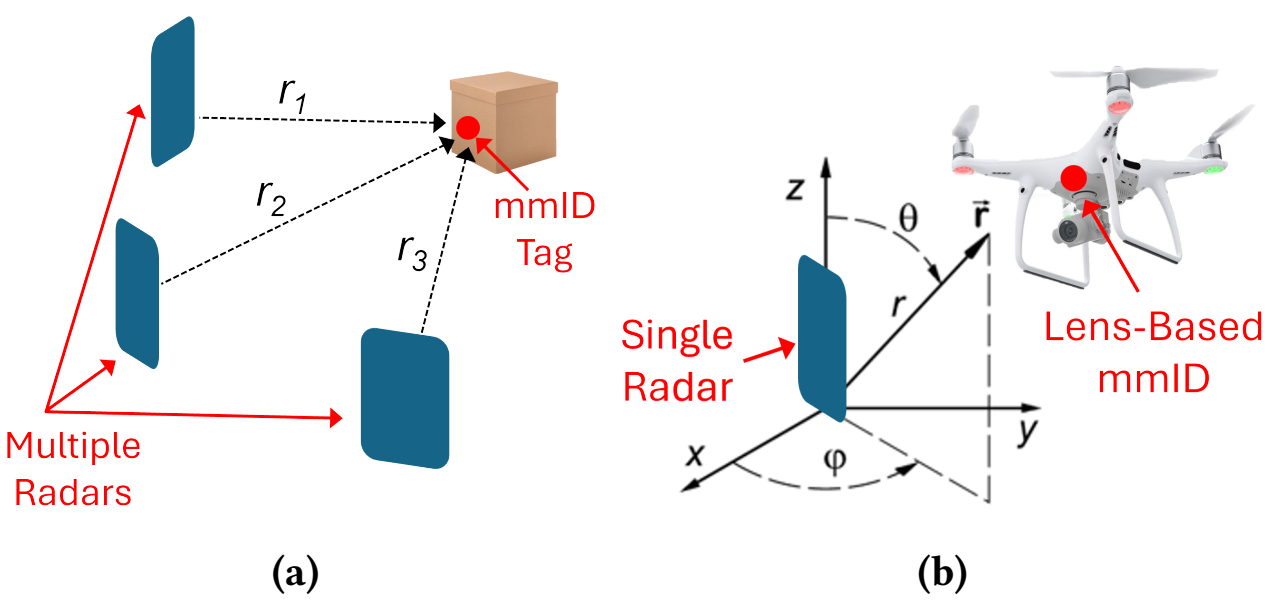

**Figure 1: mmID 3D Localization.** (a) State-of-the-art solutions rely on multiple radars to enable 3D localization of static or low motion targets, (b) DragonFly is the first to enable single radar 3D localization of highly dynamic targets.

## 2 BACKGROUND: FMCW Radar as mmID readers

The FMCW radar is the most common radar architecture, used in most commodity radar chips. As such, it is the radar of choice for low-cost mmWave backscatter (mmID or mmWave RFID) systems. In its most common mode of operation, it emits a radio frequency or mmWave signal with a linearly increasing frequency. During its propagation to and back from a target, this signal acquires phase. As the emitted frequency linearly increases, so does the phase of the received signal. After down-conversion using the Tx signal, this phase can be expressed as

$$\phi_{RX} = 2\pi \frac{2R}{c} \left( f_0 + St \right) \quad (1)$$

, where $R$ is the range to the target, $c$ the celerity of light, $f_0$ the starting frequency of the chirp, and $S$ the slope of the chirp. This linearly changing phase generates a frequency called the beat frequency and is expressed as $f_B = 2RS/c$. This makes it a powerful process, which can compress GHz of RF bandwidth into a few MHz of IF bandwidth, which can then deliver its range information using a simple Fourier Transform of the chirp's signal (which we call *fast time* or *range space*). Like all good things, this chirp cannot keep increasing forever. Once it reaches its maximum frequency, it is usually repeated with a period of $T_C$. This gives the process a maximum resolution of $\Delta R = c\Delta f_B/2S = c/2ST_C = c/2B$, where $B$ is the RF bandwidth of the radar. In a perfectly static environment, this chirp would periodically generate the same received signal. However, let us assume a target with a constant radial velocity of $v_R$. The phase received during the $n^{th}$ chirp could then be expressed as:

$$\begin{aligned}\phi_{RX,n} &= 2\pi \frac{2R + 2v_R(t + nT_C)}{c} (f_0 + St) \\ &\approx 2\pi \left( n \frac{2v_R f_0 T_C}{c} + \frac{2Rf_0}{c} + \left( f_{B,n} + \frac{2v_R f_0}{c} \right) t \right)\end{aligned} \tag{2}$$

| Systems | Single radar dimensions | 2D sampling time | Robust to speed/accel. | Tag Power Draw | Tag Coverage (azi/elev) | Single Radar 3D/2D Precision (@7m) | Max Range 90% Detection |
|---|---|---|---|---|---|---|---|
| **DragonFly** | **3D** | **3.4 ms** | **High/High** | **68 $\mu$W** | **100°/100°** | **12cm (3D)** | **50m** |
| **Millimetro [31]** | 2D | 38.4 ms - indoor 64 ms - outdoor | Medium/Low | 2.5 µW | 15°/60° (2x8 arrays) | 36 cm (2D, 3° AoA) | 37m-indoor 100m-outdoor |
| **Hawkeye [3]** | 2D | 16.7 s (8192 × 2048 @ 1 MSPS) | Low/Low | 7.7 mW (w/o baseband) | 80°/160° | 60 cm (2D, 5° AoA) | 180m |
| **R-Fiducial [5]** | 2D | 0.5-38ms | High/Medium-Low | 5 µW (w/o baseband) | 15°/120° (1x8 arrays) | 72 cm (2D, 6° AoA) | 6-25m |

**Table 1: Comparison with the state-of-the-art systems.**

Therefore, $f_B$ appears to be shifted by $2v_R/c$, which is often not detectable given the $1/T_C$ fast time resolution. However, a DFT over the dimension of the chirps ($n$, also called *slow time* or *Doppler space*) uncovers $2v_R f_0/c$—which is none other than the Doppler frequency—with a scalable resolution of $1/(NT_C)$, where $N$ is the maximum number of chirps considered. Given that we now know that the FMCW process most fluently speaks Fourier, it would be natural to modulate potential tags with variants of CWs—the question is whether to do it in *slow time* or *fast time*.

## 3 DragonFly clutter removal and 3D localization

### 3.1 Intra-chirp 2D Localization

The first challenge in using FMCW radars as mmID readers is that they are designed to track passive targets (clutter), and thus excel at filling the baseband with clutter-generated signals. In the context of mmID detection, however, these become interference. Recent approaches to isolate tags from clutter [3, 5, 31] rely on aggregating signals across many chirps and projecting them into Doppler space. This method is fragile: it yields only tenuous orthogonality between tag and clutter signals (especially in dynamic environments), requires long sampling times, and scrambles tag identities when they move. While [5] demonstrates a single-chirp edge case, it limits range to under 8 m and still suffers Doppler-induced range errors under mobility.

DragonFly overcomes these limitations with an intra-chirp process: by carefully selecting the modulation properties of both the tag and the FMCW radar, tag and clutter signals are separated in frequency to the point of guaranteed orthogonality—even in highly dynamic conditions.

In an FMCW system, range resolution is fixed by RF bandwidth (60 cm with 250 MHz), while the unambiguous maximum range is set by chirp length:$R_{MAX} = T_C f_S c/(4\Delta B)$,where $c$ is the celerity of light, and $T_C$, $f_S$, and $\Delta B$ are the chirp time, ADC sampling frequency, and RF bandwidth, respectively. Since clutter signals attenuate as $1/R^4$, $T_C$ can be chosen (with $f_S$ and $\Delta B$ largely fixed) so that clutter falls to negligible levels beyond a certain beat frequency, leaving a wide swath of the FMCW spectrum available for tags. Moreover, because Doppler shifts are given by $2vf_0/c$ (where $v$ is target velocity and $f_0$ the radar carrier frequency), selecting a sufficiently large $f_S$ ensures mobile clutter remains outside the tag's allocated band. Tags can then be modulated near $f_S$, appearing in the pristine, high-frequency portion of the baseband. The challenge then becomes: with all tag energy confined to a single chirp, how can it be localized?

A simple on–off switching of the tag's RCS at modulation frequency $f_m$ produces a distinct signature in the IF signal, enabling both detection and identification. Considering only the fundamental frequency of the square-wave modulation, the signal observed in the positive IF spectrum of the $n^{\text{th}}$ I channel (ignoring Q), with $N$ channels spaced $d_{RX}$ apart, can be expressed as:

$$\begin{aligned}IF_n(t) &\propto \exp\left(2i\pi\left(f_m + f_b\right)t + i\left(\phi_m + \phi_0 + \frac{2\pi n d_{RX}\sin\Theta}{\lambda}\right)\right) \\ &+ \exp\left(2i\pi\left(f_m - f_b\right)t + i\left(\phi_m - \phi_0 - \frac{2\pi n d_{RX}\sin\Theta}{\lambda}\right)\right)\end{aligned} \tag{3}$$

, where $\lambda$ is the wavelength at the lowest chirp frequency, $f_m$ and $f_b$ are the modulation and beat frequencies associated with the tag, $\Theta$ the azimuth AoA of the tag, and $\phi_m$ and $\phi_0$ the phases of the tag's modulation and propagation at antenna 0, respectively. A 2DFFT in both the time domain and the dimension of the Rx channels therefore uncovers two peaks, as shown in Figure 2. The frequency of the two peaks can be subtracted to determine the beat frequency (and, therefore, the range), while its azimuth AoA is determined by the location of the highest frequency peak in the angular space. It should be noted that these FFTs can simply be zero padded in either or both dimensions to interpolate and achieve higher precision and accuracy. Furthermore, the process is as computationally efficient as FFTs—which have been optimized ad nauseam and are, therefore, incredibly efficient.

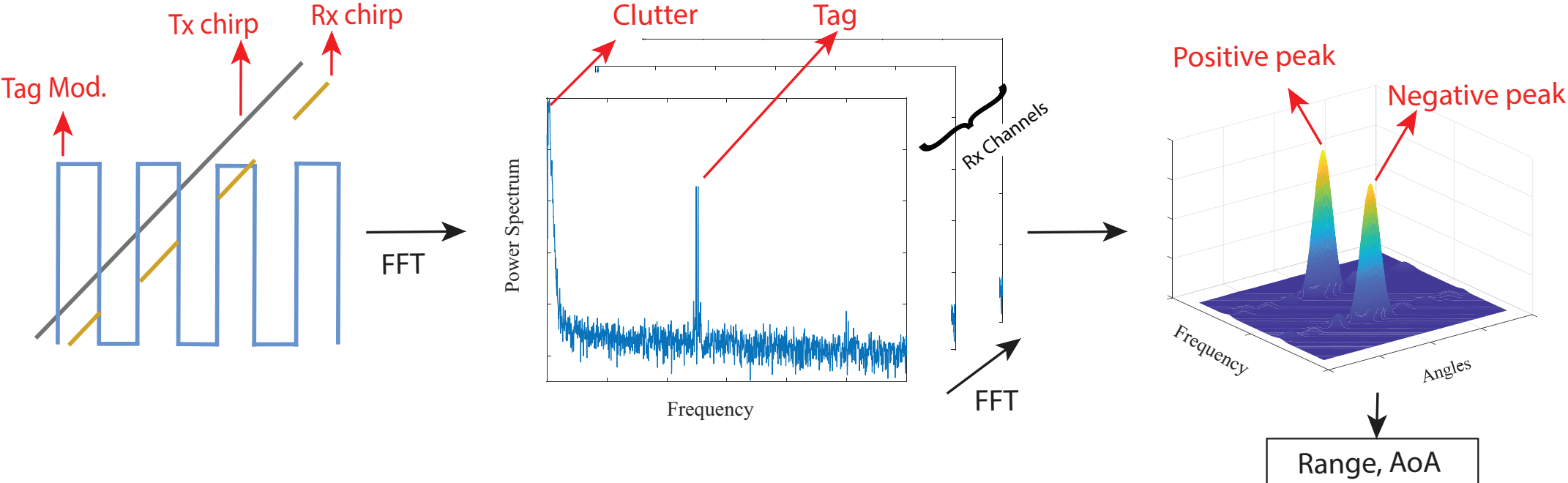

**Figure 2: 2D Localization Flowgraph.** The intra-chirp modulated signal is first Fourier transformed along fast time and then along the receiver channels, which reveals a positive $(f_m + f_b)$ and a negative $(f_m - f_b)$ peak. The frequencies and phases at these peaks are examined for precise localization of the tag.

## 3.2 Elevation Determination with Time-Divided MIMO

It is possible to use similar principles as those presented in the previous subsection to determine elevation and, thereby, finalize the single radar 3D position estimate. However, this would require the multiplication of the number of Rx channels or sacrifices in the angular resolution of the radar. In a MIMO system, multiple time divided Tx antennas can instead be utilized to offer the elevation diversity necessary to determine the elevation of the tag relative to the radar. Intra-chirp 2D localization greatly simplifies this process by allowing Tx channels to sequentially emit no more than a single-chirp at a time, thereby allowing fast switching (which, as we will see, is essential) and simple processing.

**The static tag condition** is quite simple. If we adjust Equation (3) to account for different Tx channels, the phase of each peak becomes

$$\phi_{m,k} \pm \left(\phi_0 + 2\pi\left(\frac{nd_{RX}\sin\Theta}{\lambda} + \frac{kd_{TX}\sin\Phi}{\lambda}\right)\right) \tag{4}$$

, where $k$ is the Tx antenna index and $d_{TX}$ the spacing between the Tx antennas, and $\Phi$ the elevation AoA. The now time-dependent $\phi_{m,k}$ has to be de-embedded by subtracting the phases of the two peaks but, beyond that, the signals can then be treated identically as in azimuth, where a simple FFT or linear fitting could determine the elevation. What if movement of the tag is introduced?

**The constant velocity case** becomes a bit more complicated. Let us assume that the different Tx transmissions are time divided, with a time $T_C$ between each Tx channel chirp. The phase measured on each peak then gets added a Doppler component:

$$\phi_{m,k} \pm \left(\phi_0 + 2\pi\left(\frac{nd_{RX}\sin\Theta}{\lambda} + k\left(\frac{d_{TX}\sin\Phi}{\lambda} + \frac{2v_R f_0 T_C}{c}\right)\right)\right) \tag{5}$$

, where $v_R$ is the constant radial velocity of the tag, and $f_0$ the lowest frequency of the chirp. Due to the movement of the tag and the time division of the Tx channels, the change in phase as a function of the Tx channel $(k)$ becomes dependent on both the elevation angle $\Phi$ and the radial speed $v_R$—there is a need to disambiguate what phase change is due to elevation from what is the consequence of Doppler.

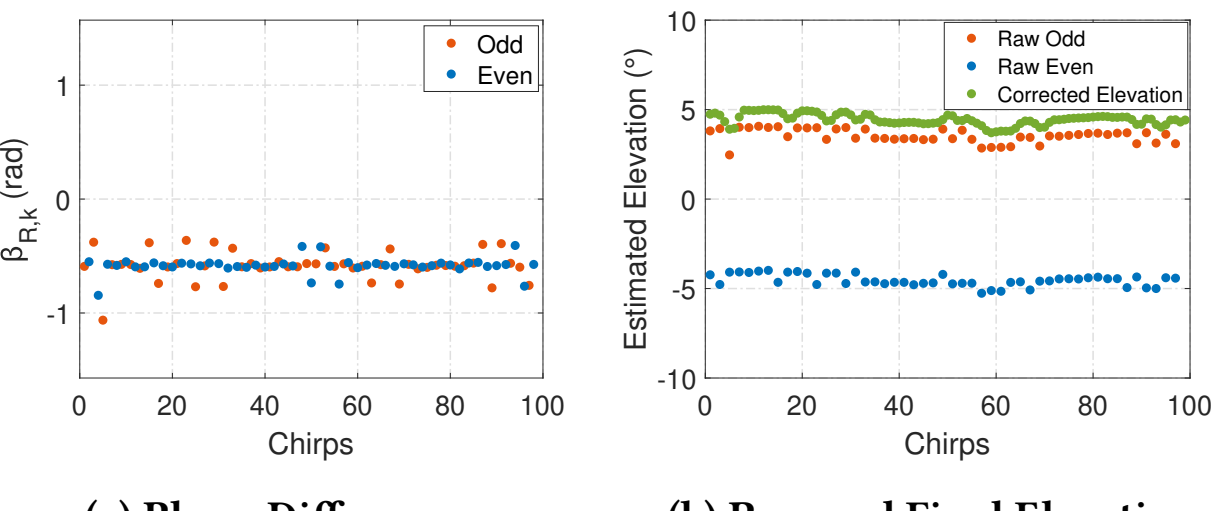

**(a) Phase Differences** **(b) Raw and Final Elevation**

**Figure 3: Constant Velocity System.** (a) $\beta_{R,k}$ for k even (blue) and odd (orange) and (b) the extracted elevations without accounting for speed (blue and orange) and using DragonFly's disambiguation method (green).

The intuition embedded in DragonFly's Doppler disambiguation process is that the periodicity of the phases outputted by the 2D analysis of each chirp with alternating channels can also be utilized to estimate the velocity of the tag, thereby allowing its subsequent elimination and the extraction of the elevation of the moving tag. We will now describe how this process, which can be extended to an arbitrary number of channels, can be implemented with two Tx channels. We start the elevation estimation procedure by calculating the radial phases $(\phi_{R,k})$ for chirp $k$, where:

$$\phi_{R,k} = 2\pi\left(\frac{nd_{RX}\sin\Theta}{\lambda} + \mathrm{mod}(k,2)\frac{d_{TX}\sin\Phi}{\lambda} + k\frac{2v_R f_0 T_C}{c}\right) \tag{6}$$

It should be noted that this phase is extracted by calculating $\left((\phi_{m,k} + \phi_{R,k}) - (\phi_{m,k} - \phi_{R,k})\right)/2$. Each of these two components being ambiguous of $2\pi$, the extracted $\phi_{R,k}$ has a $\pi$ ambiguity. Let us define $\alpha_{R,k} = (\phi_{R,k-1} - \phi_{R_{R,k}})$ and $\beta_k = (\phi_{R,k} - \phi_{R,k-2})$, for $k$ odd.

If we look at

$$\alpha_{R,k} \equiv 2\pi \left( \frac{d_{TX} \sin\Phi}{\lambda} + \frac{2v_R f_0 T_C}{c} \right) \pmod{\pi} \tag{7}$$

we realize that $\frac{2v_R f_0 T_C}{c}$ is all that stands between us and our ability to determine the elevation $\Phi$.

Now,

$$\beta_{R,k} \equiv \frac{8\pi v_R f_0 T_C}{c} \pmod{\pi} \tag{8}$$

Therefore, if one were to determine the velocity $v_R$ from $\beta_{R,k}$, one would get:

$$v_R \equiv \frac{\beta_{R,k} c}{8\pi f_0 T_C} \left( \mathrm{mod} \frac{c}{8 f_0 T_C} \right) \tag{9}$$

This estimate is highly ambiguous. For instance, for a system operating at 24 GHz with a 6.8 ms chirp time, the velocity would be ambiguous by about $0.23\,\mathrm{m\,s^{-1}}$. This velocity is not very high. However, a key insight mostly solves this issue: *Although the determined velocity is infinitely ambiguous, the fact that $\alpha_{R,k}$ is sampled twice as fast as $\beta_{R,k}$ only leaves two ambiguous phase options to select from.* This is analogous to trying to determine the bin of a tone (which might be infinitely aliased) in a Fourier transform, knowing its bin in another Fourier transform sampled half as fast as the original one: only two options are possible. Indeed,

$$\alpha_{R,k} \equiv \frac{2\pi d_{TX} \sin\Phi}{\lambda} + \frac{\beta_{R,k}}{2} + \frac{n\pi}{2} \pmod{\pi} \quad (n \in \mathbb{Z}) \tag{10}$$

**Algorithm 1** Constant Radial Velocity Condition

1: **Input:** $IF(k)$
2: **Given:** $k \in \mathbb{N}, k > 2$
3: **Raw Data Processing**
4: *Phase measurements* $\leftarrow$ *Peak selection* $\leftarrow$ *2D FFT*
5: $\phi_{R,k} \leftarrow \frac{\mathrm{diff}(\phi_{m,k}+\phi_{R,k}, \phi_{m,k}-\phi_{R,k})}{2}$
6: Record the range and estimate the direction of travel
7: **Velocity Induced Phase Estimation**
8: $\beta_{R,k} = (\phi_{R,k} - \phi_{R,k-2})$
9: **Elevation Correction**
10: **if** $\mathrm{mod}(k, 2) \neq 0$ **then**
11: $\quad \alpha_{R,k} = (\phi_{R,k-1} - \phi_{R,k})$
12: $\quad \delta_{R,k} = \alpha_{R,k} + \frac{\beta_{R,k}}{2}$
13: **else**
14: $\quad \alpha_{R,k} = (\phi_{R,k} - \phi_{R,k-1})$
15: $\quad \delta_{R,k} = \alpha_{R,k} - \frac{\beta_{R,k}}{2}$
16: **end if**
17: Unwrap the phases ($\delta_{R,k}$) and convert to Elevation

In a system with only two Txs, this entire process can be repeated with even $k$ values, remembering that the extracted value of $\sin\Phi$ has to be inverted to account for the reversed antenna order. Once the two possible options for $\alpha_{R,k}$ have been determined, two estimations of the elevation, $\Phi$, can be calculated. Knowing that the velocity is constant, any newly calculated value of $\beta_{R,k}$ can be compared to $\beta_{R,k-1}$ and picked to ensure that these are nearly equal, thereby defining a phase trajectory. Of the two possible trajectories generated by the calculation, one can then be selected based on its probability, informed by the mechanical dynamics of the mobile system that the tag is tracking and by the gains of the radar's antennas. It is typical for Tx channel antennas to be far more than half a wavelength apart. Therefore, a subsequent unwrapping of the phase trajectories can then be used to enable elevation tracking beyond the total unambiguous elevation range of the system. The phase outputs of DragonFly's process are shown in Figure 3 and its algorithm in constant velocity conditions is summarized in Algorithm 1.

**Introducing the limit: Acceleration.** Now that we have shown a path to handling velocities, no matter how high (in principle), attention should be given to practical dynamic systems, which display changing speeds—they accelerate. The system becomes significantly more chaotic when the tag undergoes acceleration. A typical set of phase plots can be seen in Figure 4, where the points appear to be scattered randomly, at least visually. In such scenarios, the radial phase disambiguation approach used under constant velocity conditions becomes ineffective. Concretely, the problem here is to choose the correct value of $\beta_{R,k}$ between the two possible values with—now that acceleration has been introduced—varying values of $\beta_{R,k}$. We may be able to assume a limited acceleration for the system and, therefore, pick $\beta_{R,k}$ values sequentially based on their proximity with their predecessor. Knowing that these two $\beta_{R,k}$ values are consistently $\pi$ apart, DragonFly's algorithm simply picks the one that is less than $\pi/2$ away from its predecessor. This accommodates all accelerations which change the measured velocity by less than

$$a_{max} = \frac{c}{16 f_0 T_C^2} \tag{11}$$

For a radar system operating at 24 GHz with a 6.8 ms chirp time, this maximum acceleration would be $16.895\,\mathrm{m\,s^{-2}}$. This is a higher acceleration than what all but the most elite race cars are capable of and should, therefore, be more than appropriate for indoor environments. *It should be noted that $1/T_C^2$ is the determining factor in this equation. A change of $T_C$ by 10x (in line with the best sampling rates reported in the state of the art) would reduce that maximum acceleration to a meager* $0.169\,\mathrm{m\,s^{-2}}$. Despite the ability of this process to handle high accelerations, it was observed experimentally that, *sporadically*, vibrations were capable of generating such accelerations. These exceptions were dealt with using the process detailed in the following subsection. The entire elevation estimation process is summarized in the flowgraph shown on Figure 4.

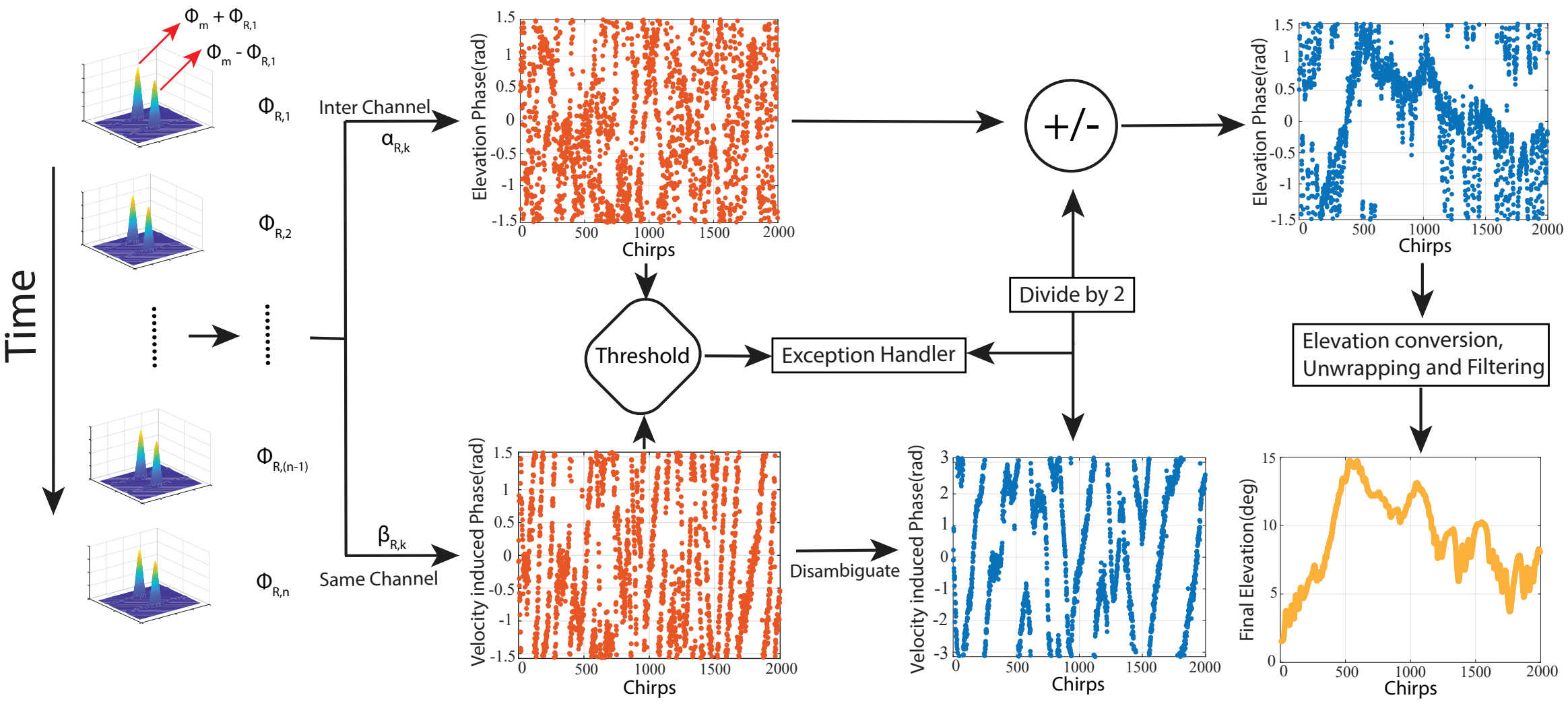

**Figure 4: DragonFly's Elevation Estimation Flowgraph.** The velocity-induced phases ($\beta_{R,k}$) and raw elevation phases ($\alpha_{R,k}$) are calculated from the radial phases derived through 2D FFT analysis of each chirp. ($\beta_{R,k}$) is then disambiguated and subsequently eliminated from ($\alpha_{R,k}$). This is used in the computation of elevation, which is unwrapped and filtered to produce the final estimate.

## 3.3 Exception Handling

It was observed in some experiments performed with a drone that there might be a few instances with large spikes in acceleration, which go beyond the maximum allowable limit. Since we are conducting a time series analysis, these outliers can distort the measurements. Here a heuristic hypothesis testing model used to handle them is described.

The intuition behind our approach is that, even at the locations of these outliers, the radial acceleration should not exhibit drastic changes and the elevation should more or less follow a predictable trajectory. Therefore, it is crucial to utilize prior information about the tag's location to resolve any ambiguities in its current velocity-induced phase. To achieve this, we first deploy two separate Kalman filters: one with an acceleration-based state transition model (STM) for handling the velocity-induced phases, which dictate the radial velocity and acceleration of the tag, and another with a constant-velocity-based STM for the elevation measurements. We then establish a threshold for the allowable acceleration of the tag. Any observed acceleration value that exceeds this threshold is deemed ambiguous and requires further testing.

In this procedure, the velocity along the radial vector ($v_r$) and along the elevation vector ($v_e$) are treated as independent random variables. Two possible velocity-induced phases, $\beta_1$ and $\beta_2$, are defined as hypotheses $H_1$ and $H_2$. Based on the respective velocities generated by the Kalman filters, we assign the conditional probabilities $P(H_1|v_r)$, $P(H_2|v_r)$, $P(H_1|v_e)$, and $P(H_2|v_e)$. These represent the likelihood of each hypothesis given the observed velocities. The individual likelihoods are then heuristically combined, and the hypothesis with the higher combined probability is selected as the final phase.

## 4 Wide Coverage High-RCS mmWave Front End

It should now be clear that DragonFly requires relatively short sampling times for each point. Otherwise, its maximum acceleration would quickly become too low for reliable use. With short sampling times, unfortunately, comes low processing gains and low Signal-to-Noise Ratios (SNRs). Therefore, it is paramount for the system's hardware to be of optimal performance. Also, given the 3D movements now afforded to the tags, these need to accommodate large coverage angles in both azimuth and elevation. Furthermore, its fully integrated cost and power consumption were to remain low, in line with the spirit of backscatter devices. To address these constraints, we propose the first fully-integrated lens-based tag that provides a wide angular coverage in both the azimuth and elevation planes, while maintaining a small form factor and a low cost. In this section, we cover the tag's main subsystems.

## 4.1 The Dielectric Lens

Retrodirectivity is an absolute necessity for any mmID system striving to achieve long ranges. Indeed, a single antenna like a patch has an aperture of only about 15 mm$^2$ at 24 GHz, which is too small to receive much of the impinging signal. Therefore, large apertures consisting of multiple antennas are necessary. However, a perfect array of a reasonable practical aperture of 10 cm$^2$ has a gain of 20 dBi, thereby only being able to cover a very narrow solid angle. It is, therefore, necessary to guarantee that the reader is always in that direction.

While Van Atta reflectarrays have been put to good use to solve this problem, their efficiencies and scalability are limited. However, one can find inspiration in the long history of optical systems and, in particular, from digital cameras. A modern camera is constituted of a lens system, whose large aperture harvests a large quantity of light before focusing it onto an array of sensors. The lens thereby generates a one-to-one map between particular directions in its far-field and its sensor's pixels. What if mmWave signals could similarly be focused by a lens onto an array, not of sensors but of independent backscatter pixels? This is what a lens-based tag achieves, as shown in Figure 6. This creates a structure that is not only very detectable but also very easily and efficiently scaled by simply changing the dimensions of the lens and extending its array made out of strictly identical backscatter pixels. Each pixel independently creates a narrow, high-gain beam through the lens. When the beam of a given pixel dies off as the angle moves away from its boresight, a neighboring pixel takes over, and so on and so forth.

Due to its simplicity and performance, a single biconvex lens made of ultra low loss Polytetrafluoroethylene (PTFE) material was selected, designed, and simulated in Ansys High-Frequency Structure Simulator (HFSS). A fundamental trade-off between the dimensions of the tag and its field of view has to be balanced, educated by the following approximation:

$$A.O.F = 2\tan^{-1}\left(\frac{h}{2F}\right) \tag{12}$$

Here, $F$, $h$, and $A.O.F$ denote the lens's focal length, the distance between the center pixel and the outermost pixel of the array, and the field of view of the lens. Once the focal length is set, we can find the radius of each convex part of the biconvex lens by using the following equation:

$$R = F * 2(\sqrt{\epsilon_r} - 1) \tag{13}$$

, where $\epsilon_r$ is the relative dielectric permittivity of the lens' material.

## 4.2 The Antenna and Switch Pixels

As we will see later, one can reap significant benefits from the cross-polarization of the signals backscattered by the tag. Therefore, simple square patches were selected, shown in Figure 7, with the two sides of each patch interconnected to ensure that the re-transmitted signal has polarization orthogonal to the incident wave. The measured $S_{11}$ of the patch antenna is shown in Figure 8 demonstrating its operation at 24 GHz. A square grid pixel arrangement was selected for simplicity to back the lens. The antenna array was modeled in HFSS (along with the tag's lens) and the combined radiation patterns of all individual antennas were simulated.

The distance between the patch antenna based elements of the backscatter array backing the lens, is one of the main design parameters. In order to assess the influence of this parameter, a single patch antenna was excited in Ansys HFSS, placed behind a model of the PTFE lens, and moved along the surface behind the lens to generate the radiation patterns shown in Figure 5. This shows all the individual but overlapping beams whose squared sum generates the total RCS of the tag, whose response is shown in Figure 15a and Figure 15b. It is apparent that the fundamental tradeoff is between the overlap level—i.e. how low in gain a beam is left to decrease before its neighbor takes over—and the complexity and cost of the array via the multiplication of array elements and switches. Quantifiably, it can be assessed that a decrease of the element spacing from the 7.35 mm chosen for our design to a more closely-packed 3.675 mm would only improve the overlap level by about 3dB, while requiring nearly 4x more elements. The maximum gain of the structure, however, remains entirely unaffected by this change, as it is solely defined by the properties of the lens and its aperture. It is important to note that it is not possible with this design to achieve a perfect focus of the lens onto the pixels of the planar pixel array regardless of angle—if the lens is perfectly focused at the center, the edges will become slightly out of focus, and vice versa.

In order to facilitate the design and operation of the patch antenna array, these were probe-fed for each polarization and their switches (Figure 9) placed on the other side of the tag. The mmWave switch is composed of high-frequency CEL CE3520K3 transistors, featuring a low operating voltage of 1.8 V, virtually null static power consumption, and an ultra-low dynamic power draw with a gate capacitance of approximately 300 fF. Its measured transmission coefficient $S_{21}$ under both biasing states is shown in Figure 10. It features two RF shorted source terminals that are connected to the drain when in zero bias, thereby connecting a shorted $\lambda/4$ stub (equivalent to an RF *open*) to the line connecting the two ports of each patch antenna and letting the signal go through. When the gate to source voltage is driven to the negative threshold voltage, the connection between the drain and source is cut off, and an open stub of $\lambda/4$ (equivalent to an RF *short*) is connected, leading to a reflection of the signal back to the port whence it came (which is not cross-polarized).

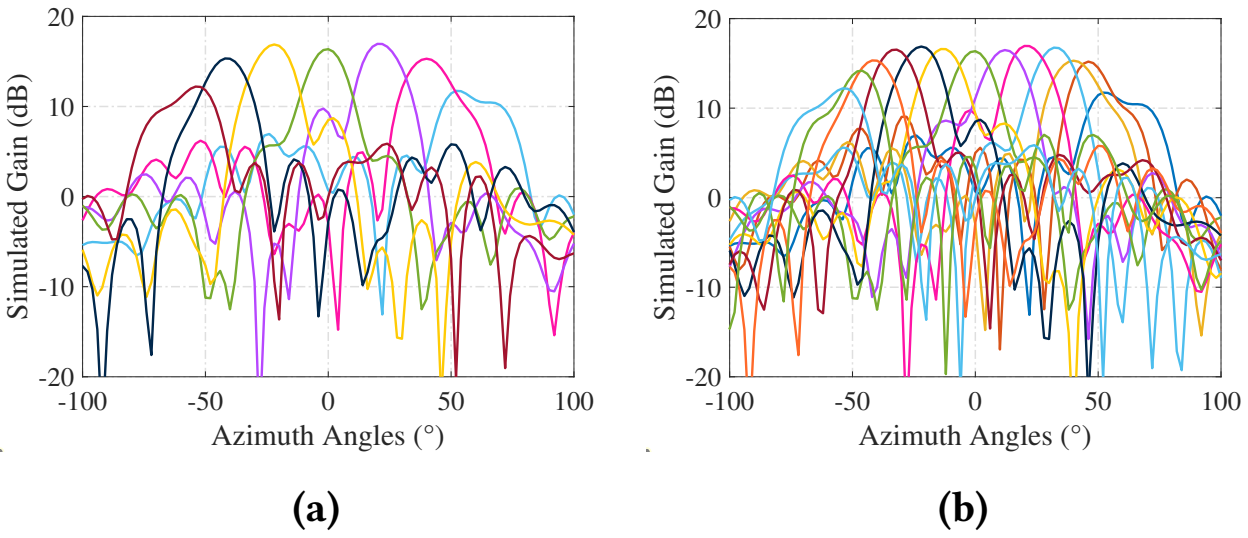

**Figure 5: Tag Gain vs Element Spacing.** Plot of the tag's simulated gain in the azimuth plane for (a) the proposed tag with a center-to-center element spacing of 7.35$mm$, (b) a tag with closely-packed antenna elements with a center-to-center spacing of 3.675$mm$.

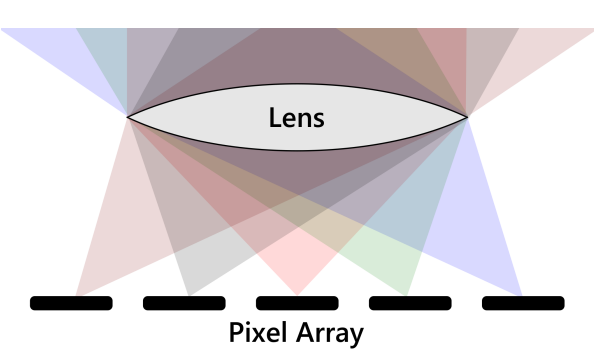

**Figure 6: Principle of Lens mmID.** The lens focuses waves coming from particular directions into individual pixels of the array.

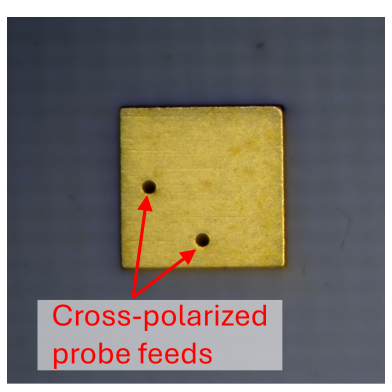

**Figure 7: Antenna Pixel.** Picture of a pixel's probe-fed cross-polarized patch.

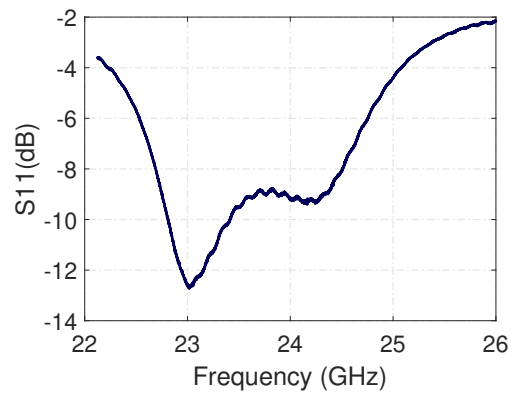

**Figure 8: Measured $S_{11}$.** Antenna measured S11 showing matching in the desired band.

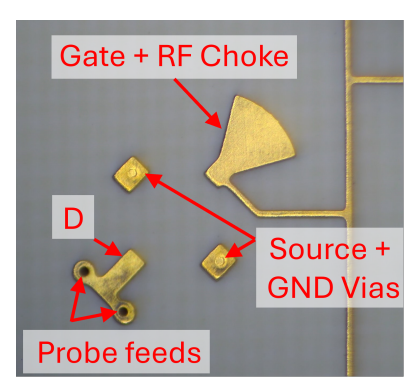

**Figure 9: Switch Pixel.** Picture of switch footprint.

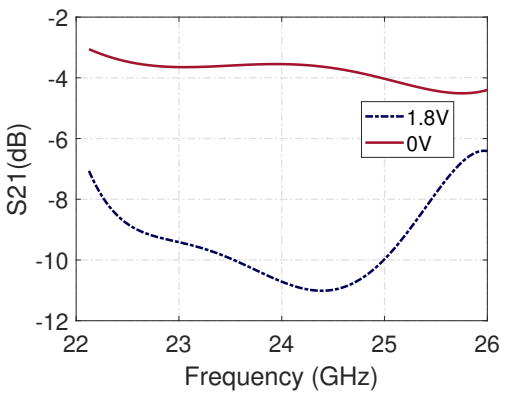

**Figure 10: Measured Switch $S_{21}$.** Transmission between the two antenna ports of the switch in 0 & 1.8 V biasing conditions.

## 5 System Implementation and Evaluation

In this section, we describe the implementation of DragonFly including fabrication of the integrated tag, and its evaluation using the radar, ground truth, and baselines.

**Integrated Lens Tag.** The proposed tag is a fully integrated system that includes a 3D PTFE lens, cross-polarized antenna elements, mmWave switches, and a baseband circuit responsible for frequency modulation. The side and bottom views of the tag are shown in Figure 11 and Figure 12, respectively. The lens was machined from bulk PTFE. Behind the lens, at a 20*mm* distance, 29 antenna elements, arranged along a 7.35 mm square grid, were built on a multi-layer Rogers 4350B 1.2*mm* thick PCB with the complete tag occupying an area of $7.46x6cm^2$. While the antennas occupy the top layer, the mmWave switches are placed on the bottom layer and connected to the antenna elements through vias. This placement was intentional to allow for a more effective placement of the antenna elements and keep the tag compact. The choice of the spacing of these elements and of their number is detailed in the previous section(Sec. 4.2). The mmWave switch is composed of high-frequency CEL CE3520K3 transistors [4]. When modulation is applied across the gate and source terminals, in the form of a square wave with amplitudes varying from 1.8V to 0V, the incoming mmWave is alternately let go from one antenna port to the other and reflected back to its original port.

The baseband circuit consists of a voltage regulator that is powered by a source that could be a battery, solar panel, or other energy harvesting devices. This voltage regulator supplies a constant voltage of 1.8V to the LTC6906 oscillator [20], configured to oscillate between 100 kHz and 600 kHz. In this work, each tag was configured with a unique frequency used to identify it.

**Power Consumption and Cost.** Each switch alone consumes a few nano watts of power, which is multiple orders of magnitude lower than what is reported in [3, 31]. The overall power consumption of the tag is dominated by the baseband oscillator, estimated to consume 12 µA at 100 kHz. The power consumption of the entire tag is measured to be 68 µW at a modulation frequency of 250 kHz. Furthermore, the tag is comprised of a single low-cost oscillator and 29 FETs with an at-scale cost of less than 5¢, thereby featuring a unit cost under 2$.

**Commercial mmWave Radar.** The radar used to interrogate the tag is an off-the-shelf cross-polarized FMCW radar from Atheraxon Inc. [1]. This radar chirps over the course of 3.4 ms and boasts 8 receive and 2 transmit channels horizontally and vertically placed $\lambda/2$ and $2\lambda$ apart, respectively—providing excellent angular coverage and resolution. It has an EIRP of 29dBm. Operating with a transmission of horizontal polarization and reception of vertical polarization, the radar is uniquely adequate for mmID applications.

**Sampling time.** The settings of the radar were uniform across all measurements: each chirp consists of 4096 points acquired over 8 channels at 1.2 Msps and zero-padded to 16384 points and 1024 points, respectively for range and azimuth interpolations. The radar is set to alternatively and regularly transmit from Tx1 and Tx2, leaving one unsampled chirp between channels to remove switching-induced dynamics, and yielding a total cycle time of 13.6 ms.

**Baselines.** We implemented two baselines:

- *Millimetro [31]:* represents a state-of-the-art system for low-power and long range accurate mmWave localization. They present a Doppler space, matched-filtering-based algorithm to localize retro-reflective Van-Atta tags.
- *Hawkeye [3]:* represents a state-of-the-art system for high tag density and long-range accurate mmWave localization. They harness the periodic nature of the FMCW radar, aggregating multiple chirps over time to distinguish and precisely range tags under clutter.

We have faithfully reproduced the authors' implementation of both systems using an ADI TinyRad mmWave radar. A co-polarized tag, modulated in On-Off Keying (OOK) at constant CW frequencies was utilized as a target for these systems.

**Drone.** We tested DragonFly with the X500-V2 drone from Holybro [39]. We attached the mmID to the drone as shown in Figure 14 and evaluated the 3D localization performance as well as trajectory error during flight.

**Ground Truth.** An OptiTrack system was used to obtain

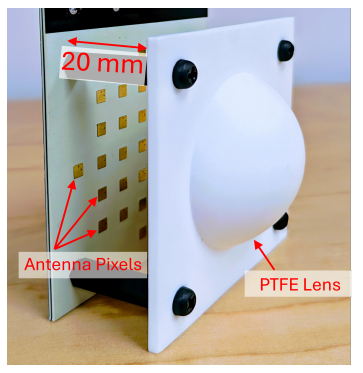

**Figure 11: Integrated Lens Tag.** Side view of the tag featuring the lens and antenna constellation.

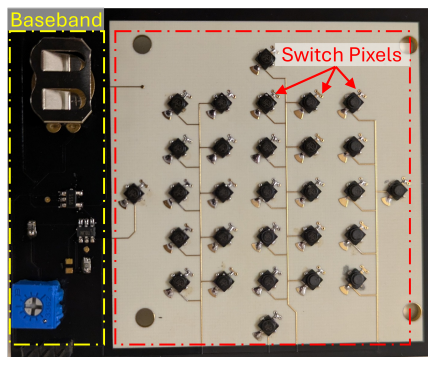

**Figure 12: Tag Bottom View.** Switch constellation form the bottom of the tag in addition to the baseband circuit.

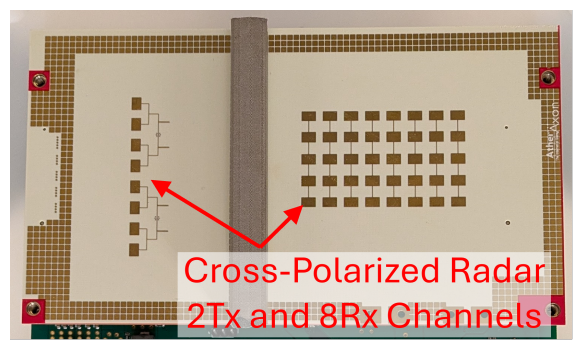

**Figure 13: mmWave Radar.** Picture of the mmWave radar used in this work featuring cross-polarized 2 Tx and 8 Rx channels.

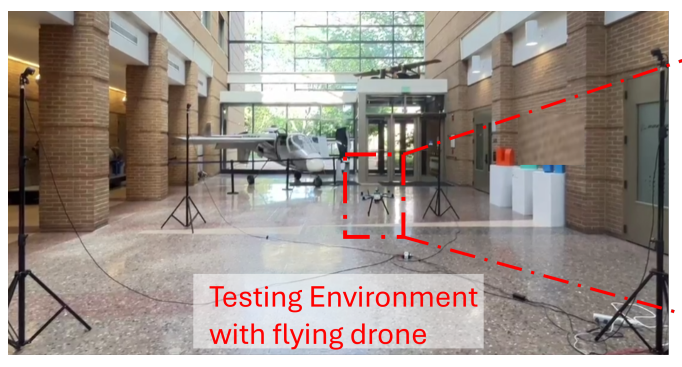

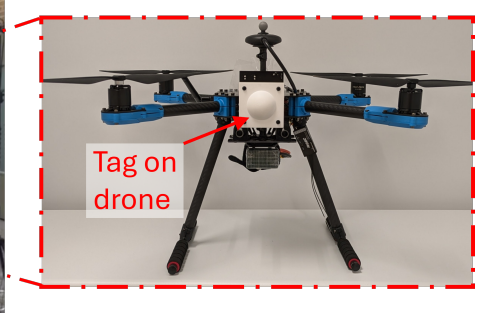

**Figure 14: Drone Experimental Setup.** Picture of environment where the 3D localization performance of the tag was assessed by attaching it to the drone and flying it in an indoor space.

ground truth location measurements [27]. The OptiTrack is an optical tracking system which consists of an array of tripod-mounted infrared cameras that can achieve millimeter-scale accuracy by relying on infrared reflective markers placed on the objects of interest.

**Evaluation Environment.** We evaluated DragonFly in different indoor environments such as large hallways, and cluttered office spaces with metallic structures. The radar was placed in the environment, while the tag was tested in varying configurations such as being attached to a drone or a toy train, held by people, or placed in static configurations to assess the 3D localization performance vs velocity, acceleration, and range. The specific experimental setup will be described in every section before showcasing the results.

## 6 Microbenchmarks

Here, we would like to understand the effectiveness of the tag's hardware implementation and design choices. We specifically focus on the PTFE lens and cross-polarization and their impact on the angular coverage and SNR, respectively.

### 6.1 Impact of Lens on Angular Coverage

In order to accurately evaluate the effectiveness of the PTFE lens on the proposed tag, we perform RCS measurements in the presence and the absence of the lens. To conduct this test, the FMCW radar data was processed and the range FFT was plotted to extract the received power at the location of the target. In the FFT analysis, the power corresponding to the target's distance is aggregated by summing the bins that contribute to the target's location. This process was repeated across both the azimuth and elevation planes at angles spanning from $-70°$ to $70°$. A cross-polarized tag with a known RCS from Atheraxon Inc. was used as a reference for the accurate normalization of the RCS measurements. The measured absolute RCS values in dBsm for both azimuth and elevation planes are shown in Figure 15. We make the following observations:

- The measured RCS in the absence of the lens in both planes displays a maximum around $0°$ and deep nulls at various angles around $10°$ and $40°$. Variations of more than 20 dB are observed across the measured angular range.
- The measured RCS in the presence of the lens in both planes also displays a maximum of around $0°$ while the drop in RCS remains manageable within 10 dB on average over $100°$ of angular coverage in each plane.

These results demonstrate that the lens enhances the coverage of the proposed tag in both azimuth and elevation planes delivering an average of 10 dB improvement across $100°$ in each plane, which translates to 5 dB enhancement in gain.

To further highlight the lens's impact on angular coverage, the RCS results were used along with the parameters and equation from Section 7.3 — to extrapolate the system's maximum theoretical range as a function of tag orientation, as shown in Figure 15a and Figure 15b. The lens-less array maintains more than half its maximum range over only about $10°$, whereas the lens extends this coverage to approximately $100°$. It is important to note that an angular coverage of $10°$ is unusable for the tracking of a drone with uncontrolled orientations. Therefore, the localization results later presented were all conducted with the lens tag.

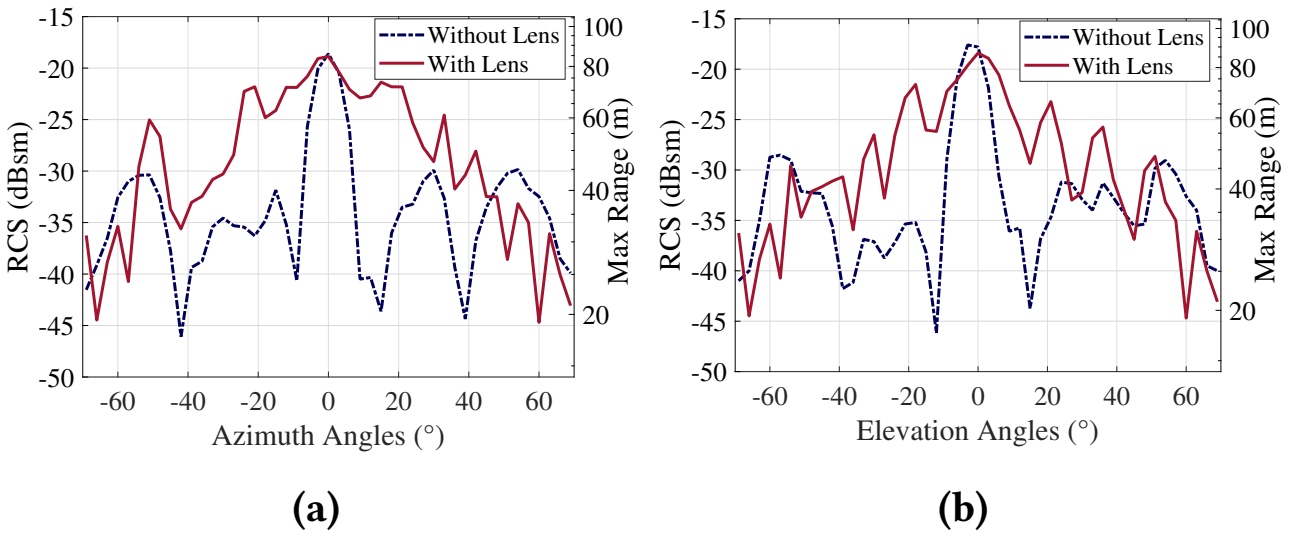

**Figure 15: Measured RCS and Maximum Theoretical Range.** Comparison of the tag's measured absolute RCS and its corresponding maximum theoretical range with and without the lens in the (a) horizontal/azimuth plane, & (b) vertical/elevation plane.

### 6.2 Impact of Cross-Polarization on Range

Commercial mmWave radars—from Analog Devices (ADI), Infineon, and Texas Instruments, etc.—are optimized to detect passive targets and, therefore, use co-polarized antennas. Our implementation utilizes Atheraxon's GreatEye radar, shown in Figure 13, which is specifically built with mmIDs in mind and, hence, relies on cross-polarized Tx and Rx channels. To study the effect of cross-polarization on the SNR performance, we implemented a co-polarized baseline using the ADI TinyRad mmWave radar—knowing that the GreatEye uses exactly the same ADI radar and ADC chipsets—and compared their responses in the same large indoor environment (a 100 m atrium), plotted in Figure 16. One can observe that the low-frequency clutter is attenuated by approximately 30 dB and that the signal of a tag placed in front of the cross-polarized radar would be entirely buried in noise otherwise. This not only leads to a more usable spectrum but also significantly lowers the spectral leakage of the large clutter scatterers into the higher frequencies. This results in the noise floor being lowered by 12 dB, which should yield a 2x in backscatter range, ceteris paribus. In this experiment, a single 4096-point chirp at a sampling of 1.2 Msps and a 45 dB power-gain ADC setting was used for both systems.

# 7 Performance Results

## 7.1 Single Radar Dynamic 3D Localization Performance

**Line of Sight (LoS) Evaluation.** We first evaluated DragonFly's 3D localization when the tag was attached to the drone as shown in Figure 14 and flown in random trajectories in an indoor space. The radar was placed at approximately 7 m away from the drone and the OptiTrack system was used to obtain the ground truth. We performed multiple experimental trials, each lasting more than 34 seconds. This allowed us to collect more than 3000 location measurements. In each trial, we performed different arbitrary movements with the drone carrying the tag. We compute the tracking error as the difference between DragonFly's location estimate and the ground truth from OptiTrack. Figure 17 plots the CDFs of the localization errors in the spherical space while Figure 18a plots them in the cartesian domain: X (red), Y (green), Z (blue) and 3D (yellow). We note the following:

- DragonFly achieved a median accuracy of $6cm$, 1.4°, and 0.38° in the range, azimuth, and elevation domains, respectively. DragonFly's 90$^{th}$ percentile is $13cm$ in range, 3.4° in azimuth, and 1.13° in elevation. This shows that DragonFly can achieve very high (sub-6 cm and sub 1.4°) median accuracy in challenging indoor environments, despite the high mobility of the drone.
- DragonFly's median accuracy in the X, Y, and Z dimensions are $3.7cm$, $8cm$, and $5.3cm$. Its 90$^{th}$ percentile is $6.4cm$ in X, $26cm$ in Y, and $11.8cm$ in Z. The total 3D error achieves a median of $12cm$, demonstrating its robustness to dynamic targets.

Figure 18 also shows the estimated trajectory by the radar compared to the ground truth. These results demonstrate DragonFly's remarkable tracking capability to enable very accurate 3D localization using a single mmWave anchor. It should be noted that the main source of localization error stems from the Y direction, which was mostly orthogonal to the radar and, therefore associated with azimuth. While the angular error in azimuth is kept mostly in check by the 8 Rx channels of the radar, the elevation errors are shrunk by limiting the unambiguous angular range of the Tx pair to ±10° and tracking the drone as it crosses the range's boundaries, using DragonFly's elevation determination algorithm.

**Non Line of Sight (NLoS) Evaluation.** In a second set of measurements, the performance of DragonFly was characterized in NLoS conditions. This was achieved using the same experimental setup as for the aforementioned LoS measurements, with the necessary and notable addition of either a piece of commercial drywall or a 2 mm thick glass window 5 cm in front of the radar. Figure 18c shows the total 3D errors for the two NLoS scenarios compared to LoS. DragonFly achieves a median 3D error of 18 cm, and 26 cm in the presence of drywall and glass, respectively, compared to 12 cm in LoS. This demonstrates the system's ability to maintain a relatively accurate 3D location of the drone under commonly encountered NLoS conditions.

## 7.2 Robustness to Speed and Acceleration

**Experimenting with Baselines.** As elaborated in Section 3, speed and acceleration can easily spell the demise of an mmID RTLS. Despite their impressive performance in certain conditions, SOTA approaches are unsuitable for most dynamic conditions. To demonstrate this, we deployed an ADI Tinyrad in an atrium also featuring a co-polarized tag modulated at 300 Hz and sampled 80x256 µs chirps. The tag was walked with—at a leisurely speed of 0.6 $\mathrm{m\,s^{-1}}$— to not only add velocity to the tag but also generate a dynamic clutter from the moving person, in what is a very gentle context, compared to DragonFly's subsequent tests. We then applied both Millimetro [31] and Hawkeye [3] to generate the plots shown in Figure 19. We can see that, both methods fundamentally rely on Doppler space, and the tag response was shifted by about 100 Hz. This is a significant issue to start with, given that tags are multiplexed and identified by using their frequencies and that a mere 1 $\mathrm{m\,s^{-1}}$ of tag velocity shifts the frequency of the tag by 160 Hz at 24 GHz, thereby quickly scrambling tag identities. Furthermore, these methods rely on various levels of averaging to improve their SNRs and detection ranges. As shown in Figure 19b, the neat peaks that Hawkeye creates in the spectrum in static environments start to spread as Doppler is introduced by some of the passive targets, thereby swallowing increasing swathes of spectrum as velocity increases. It should, finally, be noted that—even in what is *approximately* a static environment—micro-Doppler limits the noise floor around the *static* bins: a noise floor of -110 dBm is observed for a 20 ms total sampling time in Figure 19b, while a single chirp on the same system can offer -130 dBm of noise floor in the higher frequency bins with a comparably small 3.4 ms sampling time. These effects

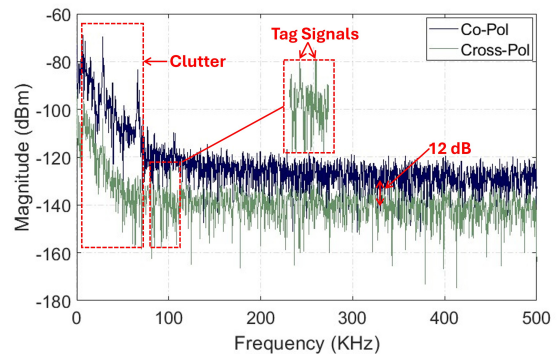

**Figure 16: Polarization Comparison.** Responses of a large indoor environment including a tag at 18 m to a co-pol. ADI Tinyrad and a cross-pol. Atheraxon radar.

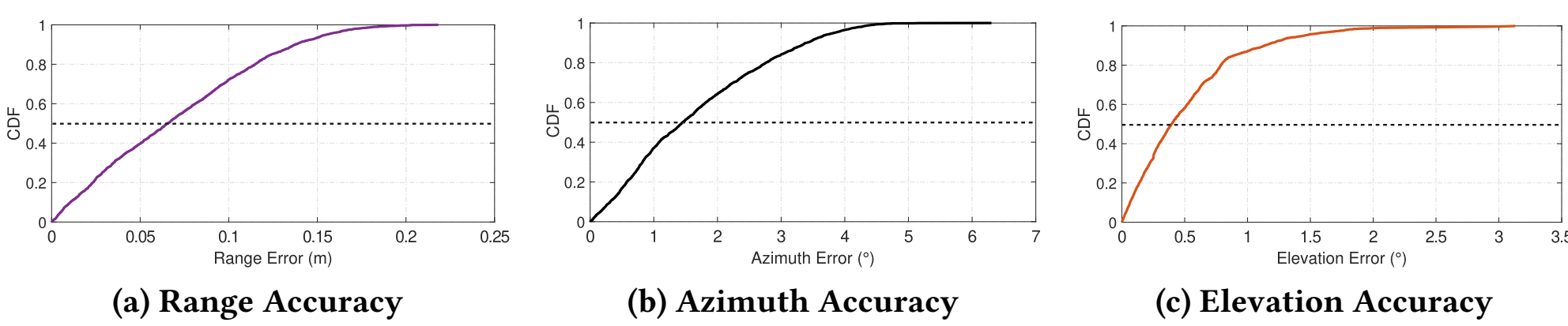

**(a) Range Accuracy** **(b) Azimuth Accuracy** **(c) Elevation Accuracy**

**Figure 17: Drone Dynamic 3D Localization.** CDF plots for DragonFly's error in range (purple), azimuth (black), and elevation (orange) during drone flight.

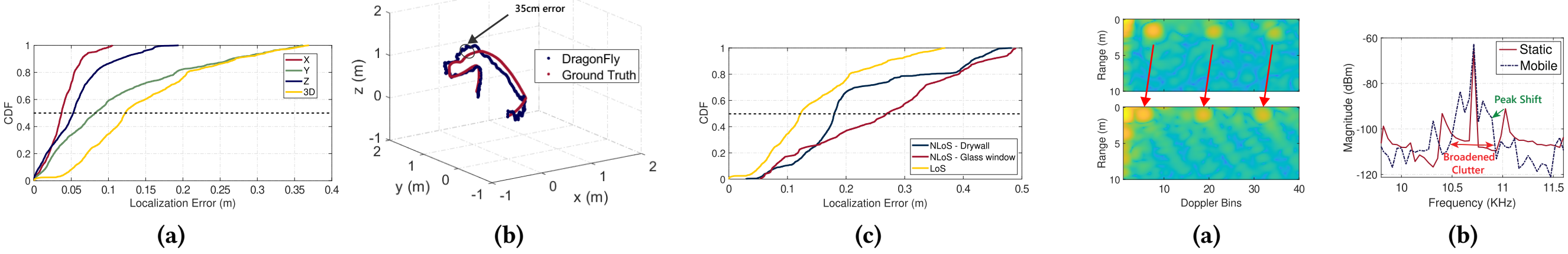

**Figure 18: Drone 3D LoS and NLoS Localization and Tracking with Single Radar.** (a) CDF plots for DragonFly's LoS 3D localization error in each of the X (red)/Y (green)/Z(blue) dimensions as well as the total 3D error (yellow), (b) trajectory tracking achieved by DragonFly in comparison to the ground truth, (c) CDF plots comparing the total 3D errors in LoS (yellow) to two NLoS scenarios: drywall (blue) and glass window (red).

**Figure 19: Behavior of SOTA Approaches in a Dynamic Environment.** (a) Range-Doppler generated by Millimetro [31] in static (top) and dynamic (bottom) conditions and (b) Spectrum outputted by Hawkeye [3].

become linearly more pronounced as the sampling times are increased to the 38.4 ms and 16.7 s used in the benchmarking of [31] and [3], respectively. Likewise, while the method introduced in [5] offers the major benefit of accommodating speed by decoupling frequency and identity, acceleration would identically distort the signal and increasingly compromise its SNR as its sampling time is increased from 0.5 ms to 38 ms (equivalent to 1 to 64 chirps).

**DragonFly vs Speed and Acceleration.** Here we evaluate DragonFly in two highly dynamic settings to assess its localization performance. In the first test, the tag was attached to a vehicle while the radar was positioned on the side of the road, as shown in Figure 20, while the vehicle drove by at increasing velocities up to 10 m s$^{-1}$ (around 22 mph). Due to the significant complexity of measuring high-speed trajectories over tens of meters with centimeter accuracy, the ground truth of this test was established by fitting a rectilinear trajectory with unspecified velocity—knowing the approximate geometry of the road relative to the radar—to the measured 3D trajectory. The results are shown in Figure 21. We can see that DragonFly's accuracies slowly degrade from 4 cm, 0.3°, and 0.5° to 25 cm, 1°, and 1.5° in range, azimuth, and elevation (respectively) as the speed of the vehicle increases up to 10 m s$^{-1}$. We believe that range and azimuth degradations can be largely attributed to increasing micro-Doppler and ground truth uncertainties at higher velocities. We also believe that these vibrations lead to more exceptions for DragonFly's process to handle and additional uncertainties on the elevation estimates. We believe that additional filtering of the data or stabilized mounting of the tag could help alleviate this imperfection. Finally, it should be noted that these azimuth results were entirely acquired in the configuration of Figure 20, where most incidence is near normal and azimuth measurements are more accurate. Therefore, their trend should be considered in relative terms before extrapolating their results to more diverse trajectories. Overall, this challenging experiment demonstrated DragonFly's robustness—and, notably, that of its Doppler disambiguation process—even at high velocities.

DragonFly's localization performance was also tested vs constant acceleration. To build this setup, we placed the tag on a toy train, as shown in Figure 22, and let it slide on a ramp of a specific inclination. The inclination angle allows us to calculate the acceleration of the tag and use it as the ground truth. Figure 23 displays the errors in range, azimuth, and elevation for acceleration up to 4 m s$^{-2}$. Here, range, azimuth, and elevation accuracies of DragonFly hardly suffered at all, while being put to their most serious test. This is a testament to DragonFly's ability to operate almost flawlessly in high acceleration conditions.

## 7.3 Operating Range

The maximum expected range of the reported system can be calculated using the radar equation and the system parameters in Table 2.

$$R = \sqrt[4]{\frac{P_t G_t G_r \lambda^2 \sigma}{(4\pi)^3 P_r}} \tag{14}$$

DragonFly's range performance was evaluated indoors, as shown in Figure 24, with the system held static on a structure.

| Parameters | System values |
|---|---|
| Transmitted Power ($P_t$) | 10dBm |
| Minimum Received Power ($P_r$) | -135dBm |
| Transmitter Gain ($G_t$) | 10dBi |
| Receiver Gain ($G_r$) | 12dBi |
| RCS ($\sigma$ at 0°) | 0.01 m$^2$ |
| Wavelength ($\lambda$) | 0.0125 m |

**Table 2: Wireless Link Parameters:** A maximum theoretical range of about 85$m$ can be achieved with the proposed system.

The measured maximum range of 55 m (Figure 25) loosely matches the theoretical limit of about 85 m derived from Equation (14). The range localization errors, also shown in Figure 25, have a median of less than 1 m up to 50 m, beyond which detection probabilities decline sharply. Notably, DragonFly achieves this long-range performance with a sampling time of only 3.4 ms, underscoring the high capability of its hardware.

## 7.4 Dynamic Multi-Tag Trajectory Tracking

In order to study DragonFly's performance in the presence of multiple tags, we relied on a simple experimental setup with two people–about 4 m in front of the radar—moving in an area covered by the OptiTrack. Each person was holding two tags (one in each hand) configured to different frequency channels and executing random trajectories simultaneously. These movements were tracked by the radar and the errors in 3D localization were measured and their results displayed in Figure 26 and Figure 27. We make the following remarks:

- DragonFly achieved a median accuracy of 4.6$cm$, 8$cm$, and 1.4$cm$ in the X, Y, and Z dimensions, respectively. DragonFly's 90$^{th}$ percentile is 10$cm$ in X, 21$cm$ in Y and 4.9$cm$ in Z. The 3D localization error has a median accuracy of 11$cm$. This shows DragonFly' robustness to multiple tags moving simultaneously.
- DragonFly simultaneously tracks the 3D positions of four tags and achieves remarkably accurate trajectories with a median 3D error of 12$cm$.

The addition of tags to the system did not degrade its performance in the least. It should be noted that the performance even improved, which can be attributed to the closer proximity to the radar with angular accuracies remaining constant. DragonFly could, in principle accommodate approximately 1500 tags simultaneously, with its current settings.

# 8 Related Work

**Active tag tracking:** Much work has been done on ranging using active low-power radio technologies such as Bluetooth Low Energy, ZigBee, and LoRa along with Received Signal Strength Indicators (RSSI) to determine sets of distances and trilaterate positions [7, 13, 14]. However, RSSI is unreliable in the presence of multipath and requires a large density of anchors/readers to generate 3D positions. These technologies are fundamentally challenged by their lack of compatibility with Time of Flight (ToF) measurements, which can be much more accurate and predictable. Modern Ultra-WideBand radios enable this capability, enabling efforts such as Polypoint [15], and Harmonium [16]. Others have additionally reported the use UWB for AoA measurements and triangulation [11, 38, 43]. While these approaches achieve low double-digit centimeter accuracies, they come at the cost of 50+ mW instant power consumption—leading to short battery lives at high positioning rates—and a multiplication of anchors (3 to 19) needed to achieve 3D localization. Additionally, WiFi is a ubiquitous standard whose modern MIMO implementation makes it suitable for AoA measurements, which has been used for localization by PinLoc [30], SpotFi [18], and Arraytrack [40]. Again, despite its appeal, the 300+ mW power consumption of WiFi radios and the need for a constellation of Access Points (APs) to enable trilateration limit these methods. Nevertheless, another approach was used by ToneTrack [41] to extract ToF from WiFi, which was then combined with AoA [34] to determine 3D positions with a single AP. However, this process remains very power-consuming for their large tracker and features a long 80 ms sampling time which is incompatible with mobility requirements.

**Backscatter tracking:** The localization of passive RFID tags has undergone extensive scrutiny and investigation [12, 19, 28, 32]. Their main limitations are range and localization accuracies. Efforts such as Tagoram [42], RF-IDraw [37] or others [24] to employ spatially diverse phase measurements to achieve cm-accurate or mm-accurate tracking total areas covered by these systems remain in the near field of their antenna systems (a few meters). Successful efforts to extend the range of backscatter systems, using semi-passive tags, have been reported [33], including for 3D single-reader localization [26]. Nevertheless, this comes at the cost of physically large and complicated readers and tags operating in bands spanning 900 MHz to 5800 MHz and with 60 ms latencies only applicable to static targets.

**mmID tracking:** MmIDs and their tracking using FMCW radars have been demonstrated [29] using active tags. Nevertheless, these consumed 10s of mW of power. After the introduction of retrodirective backscatter [17], many implementations of retrodirective mmIDs were proposed in the form of Van Atta reflectarrays [8–10, 25], Rotman lenses [6, 21], or 3D lenses [22, 23]. Their use for practical localization was demonstrated by Millimetro [31] and Hawkeye [3], achieving ranges in excess of 100m and mm-accurate localization. However, none of these reported systems are capable of single-radar 3D localization or of the reliable tracking of highly mobile targets.

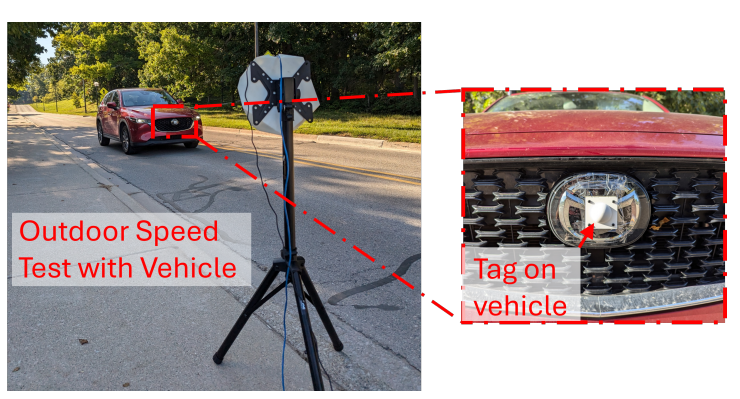

**Figure 20: Vehicle Velocity Test.** Setup used to test DragonFly's performance vs velocity up to around 22*mph* with tag attached to the vehicle and radar placed on the side of the road.

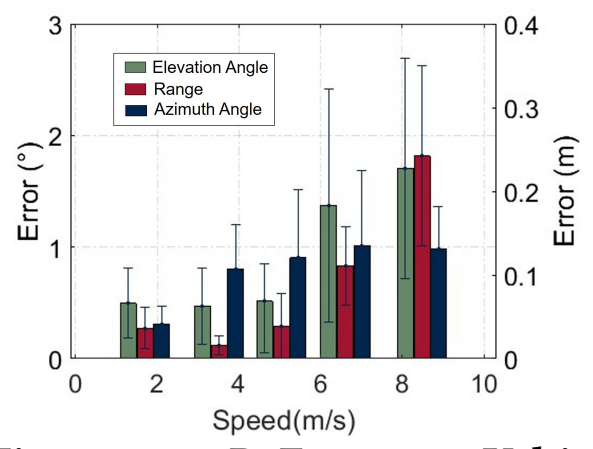

**Figure 21: 3D Errors vs Vehicle Velocity.** Measured range, azimuth, and elevation errors vs velocities up to 22*mph*.

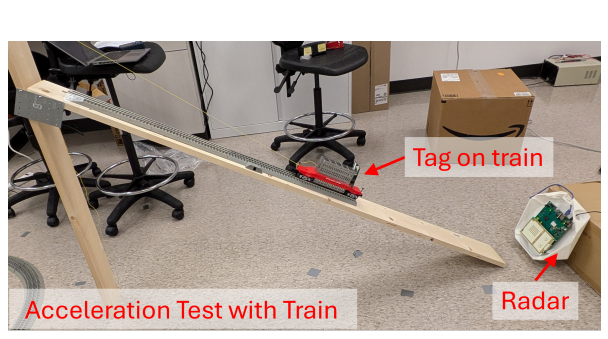

**Figure 22: Acceleration Setup.** Experimental setup involving a toy train at different inclinations to measure acceleration.

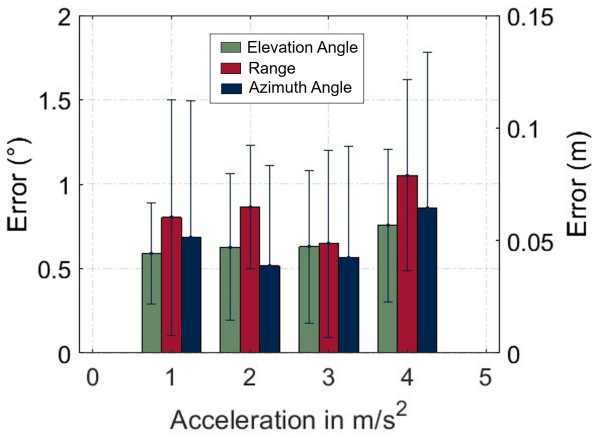

**Figure 23: 3D Errors vs Acceleration.** Measured range, azimuth, and elevation errors vs accelerations up to 4 m s$^{-2}$.

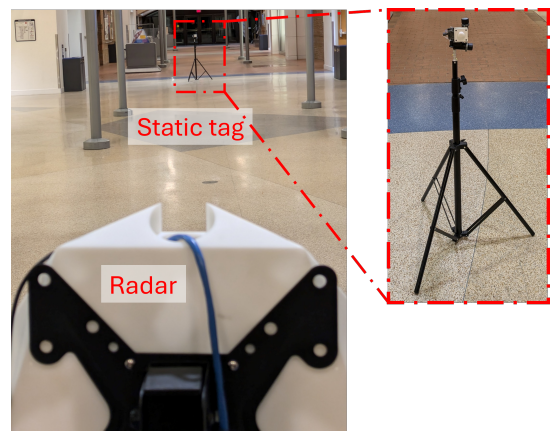

**Figure 24: Range Performance Setup.** Picture of the indoor space used to test DragonFly's performance in a static mode vs distance to the radar.

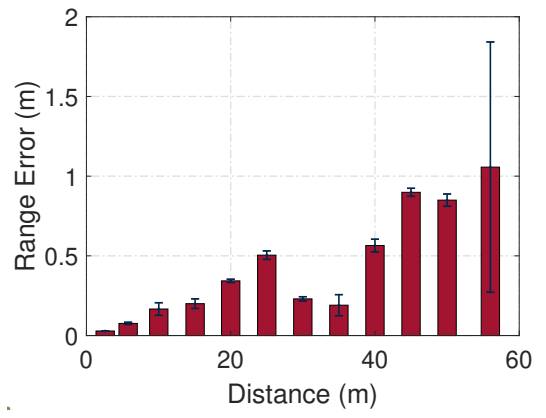

**Figure 25: Range Error vs Distance.** Range errors obtained up to a distance of 55*m*.

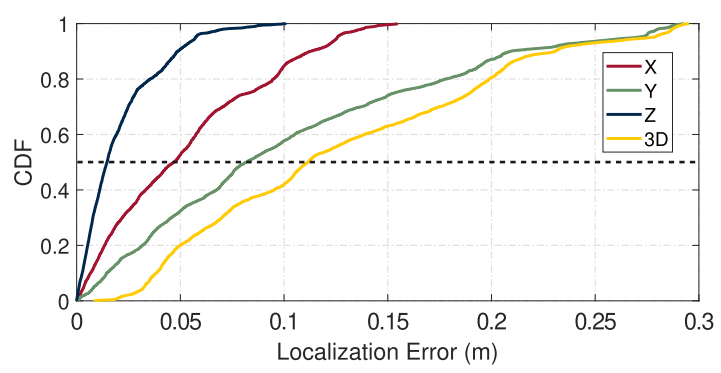

**Figure 26: Multi-Tag Simultaneous 3D Localization.** CDF plots for DragonFly's 3D localization error in each of the X/Y/Z dimensions as well as the total 3D error for 4 tags moving simultaneously.

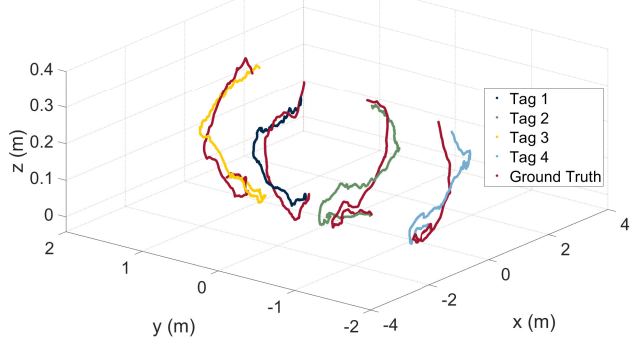

**Figure 27: Multi-Tag Simultaneous 3D Tracking.** Multi-tag trajectory tracking achieved by DragonFly in comparison to the ground truth showing a maximum error of 28*cm*.

# 9 Discussion & Conclusion

We present the first wireless system consuming microwatts of power that can be used to localize dynamic targets in 3 dimensions using a single reader/access point/anchor over ranges exceeding a few decameters. Despite this performance, one must recognize that mmID RTLSs are in their infancy. As such, we recognize the following as potentially fruitful areas of inquiry in this field:

- **Vibrations/Micro-Doppler:** Vibrations present the most pressing challenge to MIMO mmID systems, due to their high and unpredictable dynamics. Nevertheless, if deembedded properly, their effect on localization could be mitigated while also used to provide valuable information about their carrier.
- **Higher Operating Frequencies:** Frequency bands centered at 62 GHz and 78.5 GHz offer larger usable bandwidth and radiated powers. Nevertheless, mmID systems operating at these frequencies can suffer from far worse link budgets if their antenna systems are not scaled appropriately. Especially on the radar, this can lead to unworkable complexity and cost, unless MIMO is fully exploited. DragonFly sets a solid foundation for this work and offers the potential to raise the maturity and versatility of such systems.
- **Joint Communication and Sensing:** Efforts in recent mmID RTLS research have predominantly been focused on FMCW radars and CW tag modulation. Nevertheless, richer modulation schemes on both the radars and the tags have the potential to put the available wealth of mmWave spectrum to much better use by allowing the simultaneous communications of data and the localization of all radar and tags, while allowing the sensing of their passive environments.

With all these opportunities in mind, it is worth reflecting on the unequaled capabilities already offered by mmID for RTLSs and looking forward to the numerous opportunities that lie ahead.

**Acknowledgments:** We thank our shepherd and the anonymous reviewers for their valuable and constructive feedback. This work was partially supported by MTRAC Advanced Transportation and Mcity.